 \documentclass[letterpaper, 10 pt, conference]{ieeeconf}
\usepackage{multicol}
\usepackage{subcaption}
\usepackage[bookmarks=true]{hyperref}
\usepackage{algorithm}
\usepackage{algpseudocode}
\usepackage{booktabs}

\usepackage{graphicx}
\usepackage{threeparttablex}
\usepackage{amssymb}  
\usepackage{mathtools}
\def\*#1{\mathbf{#1}}
\usepackage{xcolor}

\begin{document}

\title{\textbf{DynaRetarget}: Dynamically-Feasible Retargeting using \\ Sampling-Based Trajectory Optimization}


\author{
\authorblockN{
Victor Dhédin\authorrefmark{1}, Ilyass Taouil\authorrefmark{1}, Shafeef Omar\authorrefmark{1}, Dian Yu, Kun Tao, Angela Dai, Majid Khadiv
}
\authorblockA{
Munich Institute of Robotics and Machine Intelligence
(MIRMI), Technical University of Munich (TUM), Germany.\\
Email: firstname.lastname@tum.de
}
\authorblockA{
\authorrefmark{1}Equal contribution
}
}


%

\maketitle


\begin{abstract}


In this paper, we introduce \emph{DynaRetarget}, a complete pipeline for retargeting human motions to humanoid control policies. The core component of \emph{DynaRetarget} is a novel Sampling-Based Trajectory Optimization (SBTO) framework that refines imperfect kinematic trajectories into dynamically feasible motions. SBTO incrementally advances the optimization horizon, enabling optimization over the entire trajectory for long-horizon tasks.
We validate \emph{DynaRetarget} by successfully retargeting hundreds of humanoid–object demonstrations and achieving higher success rates than the state of the art. The framework also generalizes across varying object properties, such as mass, size, and geometry, using the same tracking objective.
This ability to robustly retarget diverse demonstrations opens the door to generating large-scale synthetic datasets of humanoid loco-manipulation trajectories, addressing a major bottleneck in real-world data collection.
\end{abstract}

\IEEEpeerreviewmaketitle

\section{Introduction}
\label{sec:introduction}





Generating feasible loco-manipulation behaviors is a highly complex problem, as it requires handling the underactuation of both the robot and the manipulated object, as well as complex contact interactions. Traditional frameworks mostly relied either on gradient-based optimization to generate optimal trajectories \cite{wensing2024optimization}, or on deep reinforcement learning (RL) \cite{ha2025learning} to directly learn optimal policies. The main advantage of trajectory optimization (TO) lies in its efficiency at finding locally optimal trajectories; however, it often requires augmentation with search-based methods to enable sufficient exploration \cite{toussaint2018differentiable,ciebielski2025task}. On the other hand, while RL is very effective at generating robust behaviors, exploration remains a major challenge, frequently leading to heavy reward shaping for each individual motion.

One plausible remedy to the exploration problem is the use of demonstrations. In particular, thanks to the similarity between human and humanoid morphology, this line of research has recently attracted significant attention and has led to many impressive results \cite{zhao2025resmimicgeneralmotiontracking,pan2025spiderscalablephysicsinformeddexterous,yang2025omniretargetinteractionpreservingdatageneration}. The main idea behind these works is to retarget human motions to humanoid robots and then use the resulting trajectories as inputs to an RL policy, using domain randomization for sim-to-real transfer. In this way, the exploration problem is largely alleviated, and the RL component requires only a small set of simple reward terms that are shared across motions.

While the RL structure is largely standard among these approaches, the retargeting module differs substantially. Most methods solve a kinematic optimization problem for retargeting \cite{yang2025omniretargetinteractionpreservingdatageneration, weng2025hdmilearninginteractivehumanoid, he2025asapaligningsimulationrealworld, luo2023perpetualhumanoidcontrolrealtime}, and are therefore susceptible to artifacts such as physical and geometric inconsistencies, particularly for loco-manipulation tasks. More recently, \cite{pan2025spiderscalablephysicsinformeddexterous} proposed using sampling-based model predictive control (SBMPC) to improve retargeting quality. However, this approach solves only a short-horizon optimization problem at each iteration, making it difficult to handle long-horizon behaviors due to its inherently myopic nature.\newline

\begin{figure}
    \centering
    \includegraphics[width=0.9\linewidth]{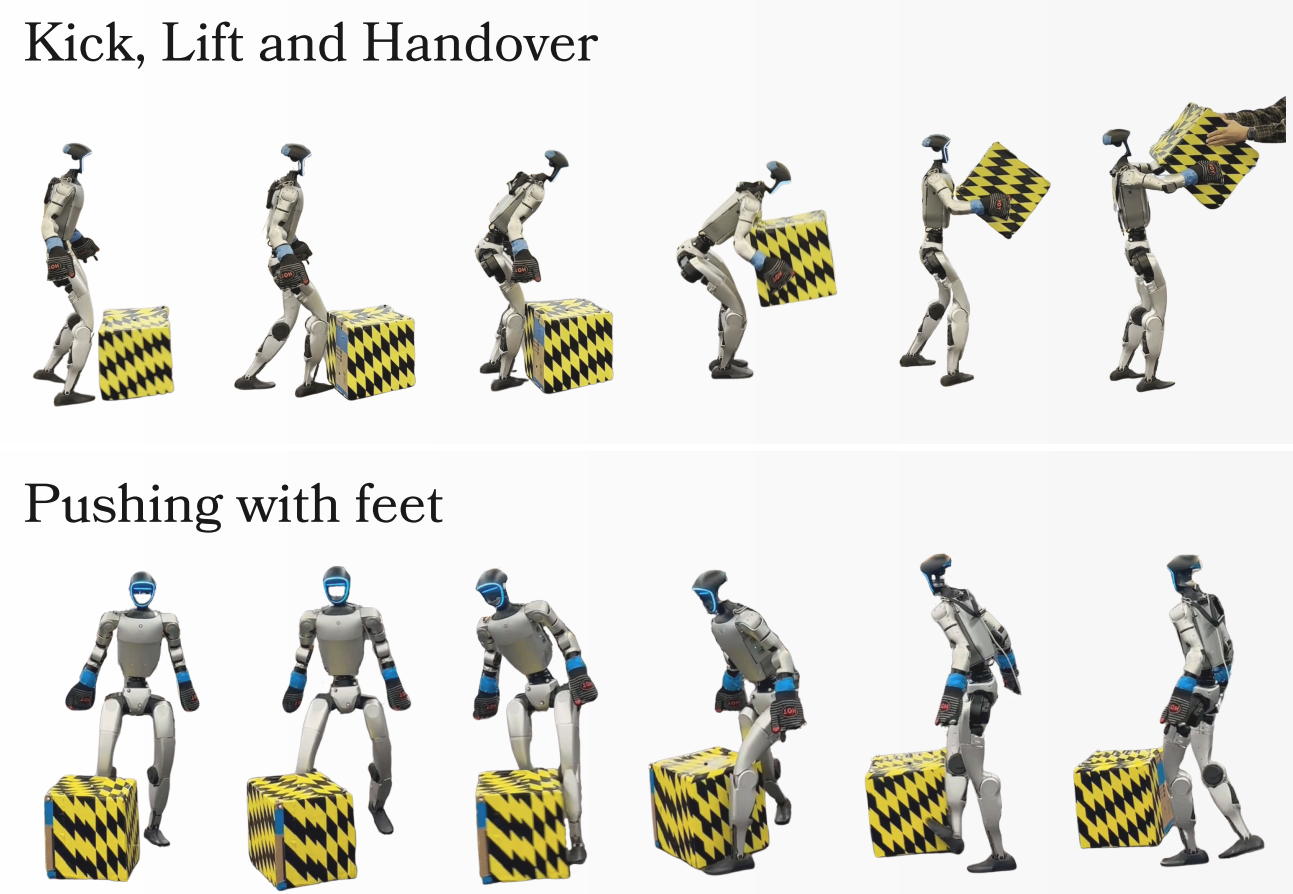}
    \vspace{-2mm}
    \caption{\textbf{Real-world humanoid loco-manipulation behaviors enabled by DynaRetarget}. Demonstrations retargeted using our framework are physically consistent and zero-shot transferable to the real robot, enabling diverse contact-rich tasks involving interactions using feet and hands, such as kicking, lifting, pushing, and object handover.}
    \label{fig:teaser}
    \vspace{-8mm}
\end{figure}

In this paper, we introduce \emph{DynaRetarget}, which combines inverse kinematic retargeting with a novel sampling-based trajectory optimization (SBTO) method that incrementally increases the optimization horizon to ultimately solve the original long-horizon problem (Fig. \ref{fig:teaser}). The generated trajectories are then fed to an RL module to learn robust tracking policies using domain randomization during training. Through extensive ablation studies and comparisons, we show that \emph{DynaRetarget} significantly improves success rates.

\begin{figure*}[h]
    \centering
    \vspace{3mm}
    \includegraphics[width=0.9\textwidth]{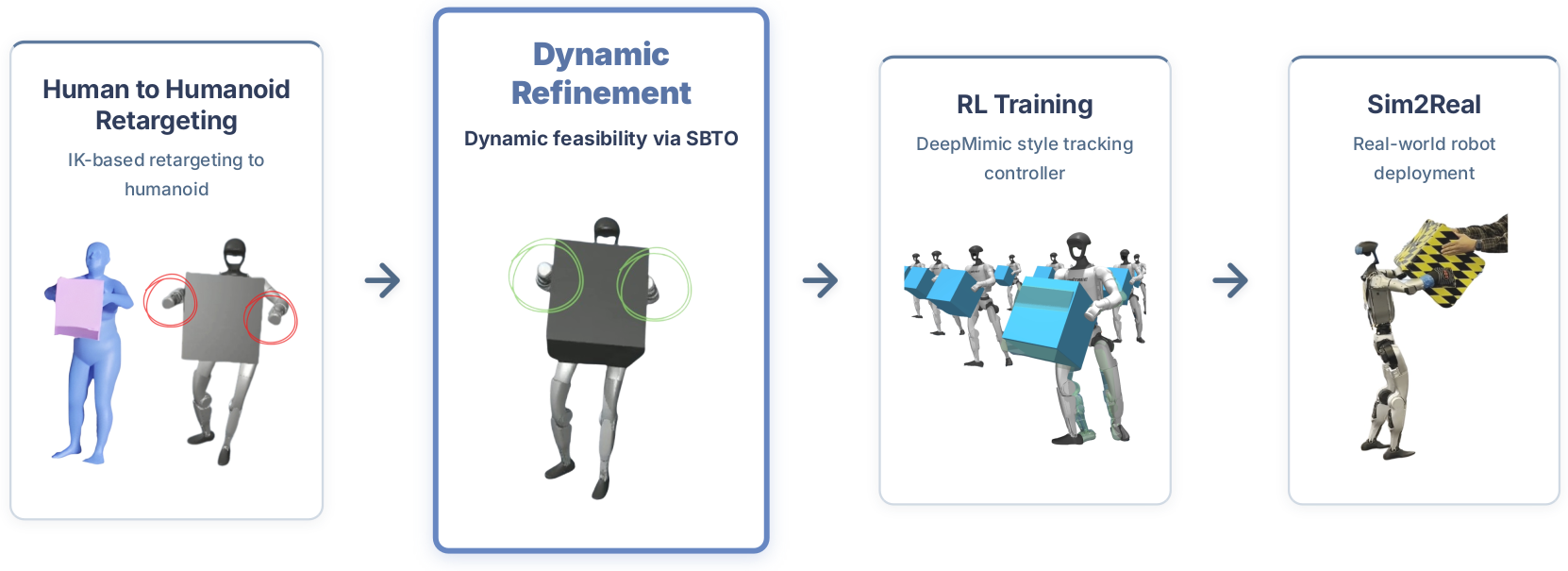}
    \vspace{-3mm}
     \caption{\textbf{DynaRetarget overview}. Given a human–object demonstration, we first perform IK-based retargeting to obtain a kinematically-feasible robot–object demonstration. Due to morphological differences between the human and the robot, this process can produce imperfections, for instance missing contacts (red circle). To address these issues, we use the kinematic trajectory as a reference for SBTO, which refines the trajectory and ensures its physical consistency, including removing missing contacts (green circle). The motion is then used to train an RL tracking policy in simulation with domain randomization. Finally, the learned policy is transferred zero-shot to our humanoid robot in the real world.}
    \label{fig:pipeline}
    \vspace{-7mm}
\end{figure*}

The main contributions of this work are as follows:
\begin{itemize}[nosep]
\item We introduce, to the best of our knowledge, the first sampling-based trajectory optimization method that retargets imperfect kinematic demonstrations into dynamically feasible humanoid loco-manipulation behaviors while considering the full problem horizon.

\item We validate our approach in simulation on hundreds of motions, demonstrating significantly higher retargeting success rates than prior methods, and show that the resulting motions improve downstream RL policy learning and transfer robustly to a real humanoid robot.
\end{itemize}

\section{Related Works}
\label{sec:related_works}

In computer graphics, motion retargeting has been extensively studied, demonstrating the effectiveness of data-driven approaches—particularly RL—for motion tracking in simulation across different morphologies using relatively simple reward structures~\cite{peng2018deepmimic, peng2020learning, peng2021amp}. Building on this line of work,~\cite{wang2023physhoiphysicsbasedimitationdynamic, xu2025intermimicuniversalwholebodycontrol} extend motion imitation to loco-manipulation by incorporating human–object demonstrations and explicitly modeling object motion and contact information in the reward function. These methods have primarily been validated in simulation.


When transferring human motions to humanoid robots, additional challenges arise from significant morphological differences, such as disparities in degrees of freedom, limb lengths, and mass distributions. PHC~\cite{luo2023perpetualhumanoidcontrolrealtime}, a retargeting method commonly used in robotics~\cite{he2025asapaligningsimulationrealworld, he2024omnih2ouniversaldexteroushumantohumanoid}, addresses this issue by selecting corresponding keypoints between the human and the humanoid robot and formulates an inverse kinematics (IK) problem based on these 3D keypoints. This is followed by an unconstrained optimization step to enforce physical consistency. However, the resulting motions often remain dynamically infeasible, exhibiting artifacts such as foot skating and penetrations. GMR~\cite{ze2025gmr} extends PHC by incorporating both keypoint positions and rotations into the IK formulation, but it still suffers from similar limitations.

Several works have explored RL-based motion tracking for humanoid robots. For example,~\cite{liao2025beyondmimicmotiontrackingversatile} introduces an RL-based framework with a reward structure similar to~\cite{peng2018deepmimic} and demonstrates the first successful deployment of highly dynamic locomotion policies on real hardware using retargeted human demonstrations. Extending this idea to loco-manipulation,~\cite{yang2025omniretargetinteractionpreservingdatageneration} relies solely on proprioceptive observations and introduces a data augmentation technique to handle variations in object shape. Similarly,~\cite{weng2025hdmilearninginteractivehumanoid} uses monocular videos to recover global human motion via~\cite{Shen_2024}, which is then used as demonstration data to train RL policies with contact-based rewards for robot–object interaction.
Although RL has proven to be a robust and effective approach for generating dynamically feasible motions from retargeted kinematic trajectories, it typically requires long training times and high-quality demonstrations. For loco-manipulation, it also requires accurate contact information which has proven to improve the RL policy performance~\cite{zhao2025resmimicgeneralmotiontracking, weng2025hdmilearninginteractivehumanoid}, but are hard to accurately obtain from kinematic retargeting.


The work most closely related to ours is~\cite{pan2025spiderscalablephysicsinformeddexterous}, which employs an SBMPC framework~\cite{xue2024fullordersamplingbasedmpctorquelevel} to improve the geometric and dynamic consistency of retargeted motions. Given a kinematically retargeted trajectory, SBMPC generates dynamically consistent motions by repeatedly optimizing control trajectories over a short horizon in a receding horizon fashion.
However, because the optimization considers only a limited horizon at each step, the method is sensitive to imperfections in the demonstration. Failures can occur during the retargeting phase, as full-horizon consistency is not explicitly enforced.

\section{Method}
\label{sec:method}

\subsection{Optimal Control with Sampling-Based Optimization}
\label{sec:sampling_based_trajectory_optimization}

Sampling-based (or \textit{zero-order}) optimization algorithms address the generic optimization problem $\min_{x \in \mathbb{R}^n} f(x)$ by only point-wise evaluating a known function $f$ to be minimized.
The gradient of $f$ is not required, which makes it popular for non-smooth and non-convex optimization problems, such as optimal control of robots for contact-rich tasks.

In optimal control, the objective is to find a control sequence $\{\mathbf{u}_0, \dots, \mathbf{u}_{T-1}\} $ minimizing a cost function $J$ while satisfying the system's dynamics  $\mathbf{x}_{t+1} = f_{dyn}(\mathbf{x}_t, \mathbf{u}_t)$:
\begin{equation}\label{eq:to_problem}
\begin{aligned}
\min_{\mathbf{u}_0, \mathbf{u}_1, \dots, \mathbf{u}_{T-1}} 
    J(\mathbf{x}_{0:T}, \mathbf{u}_{0:T-1})\\
    \text{s.t.} \quad \mathbf{x}_0 = \mathbf{x}_{ini},\;\; \mathbf{x}_{t+1} = f_{\text{dyn}}(\mathbf{x}_t, \mathbf{u}_t).
\end{aligned}
\end{equation}

$\mathbf{x}_t = (\mathbf{q}_t, \mathbf{v}_t)\in \mathbb{R}^{n_x}$ and $\mathbf{u}_t \in \mathbb{R}^{n_u}$ respectively denote the state (joint positions and velocities) and control of the system at time $t$.
To satisfy the dynamics constraint, control sequences $\mathbf{u}_{0:T-1}$ are rolled-out in a single-shooting fashion from the initial state $\mathbf{x}_0$ using a simulator (treating it as a black box), which ultimately outputs state trajectories $\mathbf{x}_{0:T}$ needed to evaluate the cost $J(\mathbf{x}_{0:T}, \mathbf{u}_{0:T-1})$.

The most commonly used algorithms in robotics are Cross-Entropy Method (CEM), a special case of Covariance Matrix Adaptation (CMA), and Model Predictive Path Integral (MPPI) \cite{jordana2025introductionzeroorderoptimizationtechniques}. In general, these approaches are used in a receding horizon fashion \cite{kurtz2024hydrax}. 
To reduce the size of the sampling space, it is common to sample \textit{interpolation knots} $\mathbf{k} \in \mathbb{R}^{K  \cdot n_u}$ instead of the full control trajectory $\mathbf{u}_{0:T-1}$. Those knots are usually equally spread in time at steps $\boldsymbol{\tau} \in \mathbb{N}^{K}$, with $\tau_0=0$ and $\tau_{K-1}=T-1$. 

\begin{algorithm}[h]
\caption{\textbf{FHTO}, Sampling-Based Fixed Horizon Trajectory Optimization}
\label{alg:zero_order_skeleton}
\begin{algorithmic}
\State \textbf{Inputs} $\boldsymbol{\mu} \in \mathbb{R}^{K \cdot n_u}, \boldsymbol{\Sigma} \in \mathbb{R}^{K \cdot n_u \times K \cdot n_u}$, $\boldsymbol{\tau} \in \mathbb{N}^{K}$, $N$ samples, $I$ iterations

\For{iteration $i = 1, 2, \dots, I$}
    \State Sample $N$ interpolation knots $\{\mathbf{k}^j\}_{j=1}^N \sim \mathcal{N}(\boldsymbol{\mu}, \boldsymbol{\Sigma})$
    \For{each sample $j$}
        \State Interpolate: $\mathbf{u}_{0:T-1}^j=\texttt{interp}(\mathbf{k}^j, \boldsymbol{\tau})$
        \State Roll-out dynamics: $\space \mathbf{x}_{t+1}^j = f_{\text{dyn}}(\mathbf{x}_t^j, \mathbf{u}_t^j)$
        \State Evaluate cost: $J^j = J(\mathbf{x}_{0:T}^j, \mathbf{u}_{0:T-1}^j)$
    \EndFor
    \State Update parameters $(\boldsymbol{\mu}, \boldsymbol{\Sigma})$ using $\{J^j\}_{j=1}^N$
    
    \footnotesize
    \hspace{-7mm}\Comment{CEM: from $N_e$ elites}
    \Comment{MPPI: exponential average weights}
    \normalsize
\EndFor
\State Return best control sequence $\mathbf{u}_{0:T-1}^*$
\end{algorithmic}
\end{algorithm}
\vspace{-5mm}

\subsection{Issues with existing SBMPC-based retargeting methods}
\label{subsec:sbmpc-issues}

Successful works in the literature that use zero-order optimization for retargeting \cite{pan2025spiderscalablephysicsinformeddexterous,kurtz2024hydrax} solve a short-horizon problem in the form of \eqref{eq:to_problem} in an MPC fashion. However, such an approach suffers from three important issues:    

\begin{enumerate}
    \item As MPC repeatedly solves a short-horizon problem, it can exhibit myopic behavior in long-horizon tasks. This effect is further exacerbated when using imperfect references that are physically inconsistent. For instance, if the contact geometry in the reference trajectory is inaccurate, the resulting short-horizon optimal plans may also fail to establish the correct contacts, ultimately leading to task failure.
    \item SBMPC-based retargeting behaves greedily: once an action is executed, the system is simulated forward, and earlier actions cannot be re-optimized. For example, if SBMPC mistakenly drops the object during the early phases of a motion, recovery is highly unlikely, as doing so would require substantial deviation from the reference, which is penalized by the tracking cost.
    %
    \item The trajectories produced by SBMPC tend to be jerky, as they are generated through feedback control. This can negatively affect both the training process and the performance of downstream RL policies.
\end{enumerate}

Considering the full horizon of the problem would avoid these pitfalls. However, humanoid loco-manipulation is a high-dimensional problem; optimizing all control variables simultaneously with a single-shooting sampling-based optimizer is therefore likely to fail, as both the number of variables and the number of local minima increase with the horizon length.
Our key observation is that the control variables toward the end of the trajectory strongly depend on those at the beginning. Updating the last control variables before the early ones are sufficiently optimized can lead to undesirable updates, making the optimization inefficient and potentially preventing convergence to desired behaviors. Based on this observation, in the next subsection, we introduce SBTO, a trajectory optimization framework that incrementally increases the optimization horizon.

\subsection{Sampling-Based Trajectory Optimization}

SBTO optimizes control variables incrementally. First, control knots $\mathbf{k}_0$ at time $\tau_0$ are optimized, then control knots $\{ \mathbf{k}_0, \mathbf{k}_1 \}$ at time $\{ \tau_0, \tau_1 \}$ (warm-starting with the previous solution $\mathbf{k}^*_0$), and so on, until all knots are optimized.
The algorithm contains two nested loops: the outer loop incrementally increases the number of decision variables being optimized, while the inner loop repeatedly refines \textit{all} the currently active variables (see Algorithm \ref{alg:zero_order_incremental_skip}).

To optimize knots $\{ \mathbf{k}_0, \dots, \mathbf{k}_k \}$, we do not perform the full horizon roll-out until $T$, but a partial roll-out until $\tau_{k}$, with $\tau_{k}$ corresponding to the time step of the last knot being optimized. Also, the optimization horizon $\tau_{k}$ grows incrementally.
At each increment $k$, this procedure is equivalent to solving the Fixed-Horizon Trajectory Optimization (FHTO) from Algorithm \ref{alg:zero_order_skeleton} with truncated parameters $\boldsymbol{\mu}_{0:\kappa}, \boldsymbol{\Sigma}_{0:\kappa, 0:\kappa}$ and knot time $\boldsymbol{\tau}_{0:k}$, where $\kappa = (k+1)n_u-1$ denotes the index of the last variable associated with knot $\mathbf{k}_k$.
Since $\tau_0=0$, the process starts at $k=1$ in practice.

Increments occur when the maximum diagonal value of the covariance matrix $\boldsymbol{\Sigma}$ is below a threshold $\sigma_{min}$, indicating that the optimization has sufficiently converged.
This adaptive criterion allows SBTO to adjust to the growing number of variables, as the convergence rate can change when more variables are being optimized. $\sigma_{min}$ is crucial for algorithm convergence; if set sufficiently large, it prevents premature convergence of $\boldsymbol{\Sigma}_{0:\kappa, 0:\kappa}$, allowing all variables in the horizon window to keep being optimized and avoiding poor local minima. If too small, the sampling distribution collapses, making it hard to escape poor local minima after incrementing. If too large, increments occur too early while variables are far from optimal, reducing the method to solving the full-horizon problem from scratch.


In practice, earlier decision variables could converge before the full horizon is reached, causing the early part of the state trajectory to eventually become effectively fixed over iterations. 
We therefore propose an improved implementation that caches these initial trajectory segments and skips their recomputation in the following roll-outs (noted \emph{SBTO\_skip}).
To do so, the algorithm identifies the latest knot $\kappa_0$ such that all earlier controls have already converged (i.e. their variances fall below a threshold $\sigma_{skip}$), and optimizes all knots between $\kappa_0$ and the last knot of the current increment $\kappa$.
This can reduce computation quite significantly for longer-horizon problems, as it is not required to roll out from $\tau_0$ at each step. This improvement can be seen in Algorithm \ref{alg:zero_order_incremental_skip}.

Note that SBTO is tailored for problems having a dense cost $J$ where even a short-horizon window provides a meaningful estimate of the optimal solution. Retargeting tasks satisfy this property, which motivates our evaluation of the proposed method in this context.

\begin{algorithm}[h!]
\caption{\textbf{SBTO} and \textbf{SBTO\_skip}}
\label{alg:zero_order_incremental_skip}
\begin{algorithmic}

\State \textbf{Inputs} $\boldsymbol{\mu} \in \mathbb{R}^{K \cdot n_u}, \boldsymbol{\Sigma} \in \mathbb{R}^{K \cdot n_u \times K \cdot n_u}$, $\boldsymbol{\tau} \in \mathbb{N}^{K}$, $N$ samples, $I=1$ iteration, $\sigma_{\min}$, $\sigma_{\text{skip}}$

\For{$k = 1, 2, \dots, K-1$}

    \State $\kappa \gets (k+1)n_u - 1$

    \While{$\max(\mathrm{diag}(\boldsymbol{\Sigma}_{0:\kappa,0:\kappa})) > \sigma_{\min}$}
        
        \If{allow\_skip}
            \small
            \State $k_0 \gets \max \left\{ j < k \;\middle|\; \mathrm{diag}(\boldsymbol{\Sigma}_{0:j\cdot n_u,0:j \cdot n_u}) < \sigma_{\text{skip}} \right\}$
            \normalsize
            
            \Comment{All variables before $k_0$ have converged}
        \Else
            \State $k_0 \gets 0$
        \EndIf
        
        \State $\kappa_0 \gets k_0 \cdot n_u$

        \State $\mathbf{u}_{\tau_{k_0}:\tau_k} \gets \text{FHTO}(
        \boldsymbol{\mu}_{\kappa_0:\kappa}, \boldsymbol{\Sigma}_{\kappa_0:\kappa,\kappa_0:\kappa},
        \boldsymbol{\tau}_{k_0:k}, N, I)$
        
        \Comment{Skip roll-out before $\tau_{k_0}$}

    \EndWhile

\EndFor

\State \Return $\mathbf{u}_{0:T-1}$

\end{algorithmic}
\end{algorithm}
\vspace{-5mm}


\section{Evaluation}
\label{sec:evaluation}

We evaluate SBTO on a trajectory refinement task using reference motions from the OmniRetarget dataset \cite{yang2025omniretargetinteractionpreservingdatageneration}. As input, we use kinematically retargeted trajectories from this dataset; in fact, SBTO performs \textit{dynamic refinement}, correcting kinematically imperfect trajectories to produce dynamically feasible whole-body motions. The dataset contains hundreds of motions of a G1 humanoid robot interacting with a box, including pick-and-place, kicking, and pushing or dragging motions. Many of these trajectories exhibit missing contacts, penetrations, or discontinuities, making them challenging to refine.


In subsection \ref{sec:implementation_details}, we provide implementation details of SBTO. Subsection \ref{sec:perf_eval} compares SBTO’s performance with a state-of-the-art SBMPC. In subsection \ref{sec:alg_analysis}, we analyze the optimization process to highlight SBTO’s key properties. Section \ref{sec:demonstration_augmentation} demonstrates that SBTO can adapt to objects with properties (shape, etc.) different from the original demonstrations. Finally, section \ref{sec:rl_eval} shows that the quality of SBTO’s output trajectories benefits the training and deployment of RL tracking policies.

\subsection{Implementation details}
\label{sec:implementation_details}

We implemented SBTO using the MuJoCo simulator \cite{todorov2012mujoco} and leveraged its newly introduced \href{https://mujoco.readthedocs.io/en/stable/python.html#rollout}{\texttt{rollout}} function to perform parallel roll-outs. We used a simulation timestep of $\Delta t=0.01$ s and considered the full collision model of the robot. The time interval between knots is independent of the reference and is set to $0.25$ s. The control sequence $\boldsymbol{u}_{0:T}$ corresponds to a PD target trajectory.

In all experiments, we use CEM to update the sampling distribution. Following \cite{pinneri2020sampleefficientcrossentropymethodrealtime}, we additionally retain a subset of elite samples across iterations ($N_{keep} = \lceil \rho_{k}\rho_e N \rceil$) and apply an exponentially weighted moving average (EWMA) with momentum parameters $\alpha_\mu$ and $\alpha_\Sigma$ to prevent premature shrinking of the distribution \cite{BOTEV201335}. The initial mean of the distribution is set to the joint positions at each knot time step from the reference, $\boldsymbol{\mu} = \mathbf{q}^{ref}_{0:T}[\boldsymbol{\tau}]$, while the initial covariance is set to $\boldsymbol{\Sigma} = \sigma_0^2 \mathbf{I}_{K \cdot n_u}$. We observed that considering the full covariance matrix improved convergence. Hyperparameter values are reported in Table~\ref{tab:cem_params}.

\begin{table}[h]
    \centering
    \caption{CEM and SBTO hyperparameters}
    \vspace{-2mm}
    \label{tab:cem_params}
    \begin{tabular}{l c}
    \toprule
    Parameters & Value \\
    \midrule
    Number of samples $N$ & 1024 \\
    Elite set proportion $\rho_e$ & 0.03 \\
    Keep elites proportion $\rho_k$ & 0.04 \\
    Mean momentum $\alpha_\mu$ & 0.95 \\
    Covariance momentum $\alpha_\Sigma$ & 0.2 \\
    Initial std. $\sigma_0$ & 0.25 \\
    Increment threshold $\sigma_{min}$ & 0.055 \\
    Skip threshold $\sigma_{skip}$ & $10^{-4}$ \\
    \bottomrule
    \end{tabular}
\end{table}

The cost function penalizes deviations in the state position $\mathbf{q}_{0:T}$ and velocity $\mathbf{v}_{0:T}$. Additional task-space terms enforce tracking of desired torso, foot, and hand poses. Contact-related terms discourage undesired collisions. The cost weights are summarized in Table \ref{tab:SBTO_cost}.

\begin{table}[h]
\centering
\caption{Cost terms and corresponding weights. $\ominus{}$ denotes the subtraction between quaternions in the tangent space. Collision costs are computed as the number of collision events, which can be easily obtained with MuJoCo.}
\vspace{-2mm}
\label{tab:SBTO_cost}
\small
\begin{tabular}{l l c}
\toprule
\textbf{Cost term} & \textbf{Equation} & \textbf{Weight} \\
\midrule
\multicolumn{3}{l}{\textit{Motion Tracking}} \\
\midrule
Joint position
& $\|\mathbf{q}_{act} - \mathbf{q}_{act}^{\text{ref}}\|^2$ 
& 0.25 \\

Joint velocity
& $\|\mathbf{v}_{act} - \mathbf{v}_{act}^{\text{ref}}\|^2$ 
& 0.01 \\

Base position
& $\|{}^{W}\mathbf{p}_{\text{base}} - {}^{W}\mathbf{p}_{\text{base}}^{\text{ref}}\|^2$ 
& 5.0 \\

Base orientation
& ${}^{W}\mathbf{q}_{\text{base}} \ominus {}^{W}\mathbf{q}_{\text{base}}^{\text{ref}}$ 
& 1.0 \\

Object position 
& $\|{}^{W}\mathbf{p}_{\text{object}} - {}^{W}\mathbf{p}_{\text{object}}^{\text{ref}}\|^2$ 
& 40.0 \\

Object orientation 
& ${}^{W}\mathbf{q}_{\text{object}} \ominus {}^{W}\mathbf{q}_{\text{object}}^{\text{ref}}$ 
& 4.0 \\

Object linear vel. 
& $\|{}^{W}\mathbf{v}_{\text{object}} - {}^{W}\mathbf{v}_{\text{object}}^{\text{ref}}\|^2$ 
& 0.2 \\

\midrule
\multicolumn{3}{l}{\textit{Motion Tracking (task space)}} \\
\midrule

Torso position
& $\|{}^{W}\mathbf{p}_{\text{torso}} - {}^{W}\mathbf{p}_{\text{torso}}^{\text{ref}}\|^2$ 
& 30.0 \\

Torso orientation
& ${}^{W}\mathbf{q}_{\text{torso}} \ominus {}^{W}\mathbf{q}_{\text{torso}}^{\text{ref}}$ 
& 3.0 \\

Torso linear vel.
& $\|{}^{W}\mathbf{v}_{\text{torso}} - {}^{W}\mathbf{v}_{\text{torso}}^{\text{ref}}\|^2$ 
& 0.3 \\

Torso angular vel.
& $\|{}^{W}\mathbf{w}_{\text{torso}} - {}^{W}\mathbf{w}_{\text{torso}}^{\text{ref}}\|^2$ 
& 0.1 \\

Foot position
& $\sum_{i \in \mathcal{F}} \|{}^{W}\mathbf{p}_{\text{foot}_i} - {}^{W}\mathbf{p}_{\text{foot}_i}^{\text{ref}}\|^2$ 
& 10.0 \\

Hand position
& $\sum_{j \in \mathcal{H}} \|{}^{W}\mathbf{p}_{\text{hand}_j} - {}^{W}\mathbf{p}_{\text{hand}_j}^{\text{ref}}\|^2$ 
& 5.0 \\

\midrule
\multicolumn{3}{l}{\textit{Regularization}} \\
\midrule

Robot--obj. collision 
& $\sum_{c \in \mathcal{C}_{\text{ro}}} 1$ 
& 2.0 \\

Self-collision 
& $\sum_{c \in \mathcal{C}_{\text{self}}} 1$ 
& 1.0 \\
\hline
\end{tabular}
\end{table}

\subsection{Performance evaluation}
\label{sec:perf_eval}

We evaluate SBTO on the OmniRetarget dataset \cite{yang2025omniretargetinteractionpreservingdatageneration} (285 motions in total).
SPIDER \cite{pan2025spiderscalablephysicsinformeddexterous}, a recently released SBMPC that achieves state-of-the-art results on many dynamic refinement tasks, serves as our baseline.
In SPIDER, the cost terms are based solely on the configurations  $\boldsymbol{q}_{0:T}$ (and not their velocities). For a fair comparison, we also evaluate a variant of SBTO using a similar configuration - considering only the terms described in the first section of Table \ref{tab:SBTO_cost} and omitting the velocity terms — referred to as SBTO\_pos.
We compare the methods using three metrics: algorithm success rate, computational efficiency, and smoothness of the resulting trajectories (all described below). The results are reported in Table \ref{tab:algo_comparison}.

We consider the refinement successful when the object trajectory has an average position error $E_{\mathrm{pos}}<10$cm and an average rotation error $E_{\mathrm{rot}} < 25^\circ$. The error terms are defined as below:
\begin{align}
    E_{\mathrm{pos}} 
    &= \frac{1}{T} \sum_{t=1}^{T} \left\| \mathbf{p}_{\mathrm{obj}, t} - \mathbf{p}_{\mathrm{obj}, t}^{\mathrm{ref}} \right\|_2 \\
    E_{\mathrm{rot}} 
    &= \frac{180}{\pi}  \frac{1}{T} \sum_{t=1}^{T}
    \arccos\!\bigl( 2 \langle \mathbf{q}_{\mathrm{obj},t}, \mathbf{q}_{\mathrm{obj},t}^{\mathrm{ref}} \rangle^2 - 1 \bigr)
\end{align}
We define computational efficiency $\eta_{\text{eff}}$ as the total number of simulation steps required in the optimization, divided by the duration of the reference. This makes the metric independent of the machine or simulator, providing a more meaningful measure than one based on compute time. For SBTO, the computational cost depends on the number of knots being optimized at iteration $i$, denoted by $k(i)$, as can be seen below:
\begin{equation}
\eta_{\text{eff}} \;=\; \frac{N_{\text{sim}}}{T \Delta t} \stackrel{SBTO}{=} \frac{N}{T \Delta t}  \sum_{i\in \mathcal{I}} \tau_{k(i)}
\end{equation}
Finally, we define the trajectory smoothness $S$ as the sum of accelerations of all \textit{actuated} joints (obtained by finite differencing) over the full trajectory. For better interpretability, we normalize the result by the trajectory smoothness of its corresponding reference $\tilde{S} \;=\; \frac{S}{S_{\text{ref}}}$.
\begin{equation}
S \;=\; \sum_{t=2}^{T-1} \left\| \ddot{\mathbf{q}}_{t} \right\|_1,
\quad \text{with} \quad 
\ddot{\mathbf{q}}_{t} = \frac{\mathbf{q}_{t+1} - 2\mathbf{q}_{t} + \mathbf{q}_{t-1}}{\Delta t^2}
\end{equation}      

\begin{table*}[h]
\centering
\caption{Algorithm performance comparison. The computational efficiency and smoothness are averaged over the successful trajectories only. For the compute, we provide absolute and relative values (separated by |).}\vspace{-2mm}
\label{tab:algo_comparison}
\begin{tabular}{lccccc}
\toprule
\textbf{Algorithm} & \textbf{Success} (\%)  $\uparrow$ &  \textbf{Smoothness} $\downarrow$ & \textbf{Compute $\eta_{\text{eff}}$} $\downarrow$ & $E_{pos}$ (m) $\downarrow$ & $E_{rot}$ (rad) $\downarrow$\\
\midrule
SBTO\_skip & \textbf{76.8} & \textbf{1.41} & \textbf{1.18e7}|0.96 & \textbf{0.11}$\pm$0.17 & \textbf{0.32}$\pm$0.35 \\
SBTO      & 74.6 & 1.7 & 4.06e7|3.3 & 0.15$\pm$0.28 & 0.36$\pm$0.44 \\
SBTO\_pos  & 62.1 & 2.7 & 4.46e7|3.6 & 0.24$\pm$0.37 & 0.42$\pm$0.45 \\
SPIDER    & 37.9 & 3.4 & 1.23e7|1. & 0.33$\pm$0.28 & 0.55$\pm$0.42 \\
\bottomrule
\end{tabular}
\vspace{-5mm}
\end{table*}

As summarized in Table \ref{tab:algo_comparison}, SBTO outperforms SPIDER, achieving nearly twice the success rate and producing smoother refined trajectories. Even SBTO\_pos shows a clear improvement over SPIDER, highlighting the algorithmic advantages of SBTO. While SBTO is computationally expensive, the SBTO\_skip variant reduces the computational cost by a factor of three, making it cheaper than SPIDER, while also achieving slightly better performance.
In practice, the refinement process takes roughly $20$ second per second of refined motion on a 112-core Intel(R) Xeon(R) Platinum 8480+ CPU.
Failure cases typically occur when references are of poor quality, especially if they include sudden changes in hand–object contact or abrupt flips in object orientation.

\subsection{Algorithm analysis}
\label{sec:alg_analysis}

In this section, we analyze SBTO’s optimization process to highlight its core features. We identify two key properties that likely explain its superiority over both Fixed-Horizon TO (FHTO) and SBMPC.
\begin{itemize}
    \item SBTO incrementally optimizes the controls, warm-starting larger-horizon problems from shorter ones up to the full horizon. This approach mitigates the convergence and instability issues observed in FHTO.
    \item SBTO optimizes decision variables over a horizon far longer than the SBMPC horizon, overcoming its inherent short-sightedness.
\end{itemize}


\begin{figure}[h!]
    \centering

    \subfloat[SBTO, $k=3$]{
        \includegraphics[width=0.3\linewidth, trim={21cm 12cm 23cm 7cm},clip]{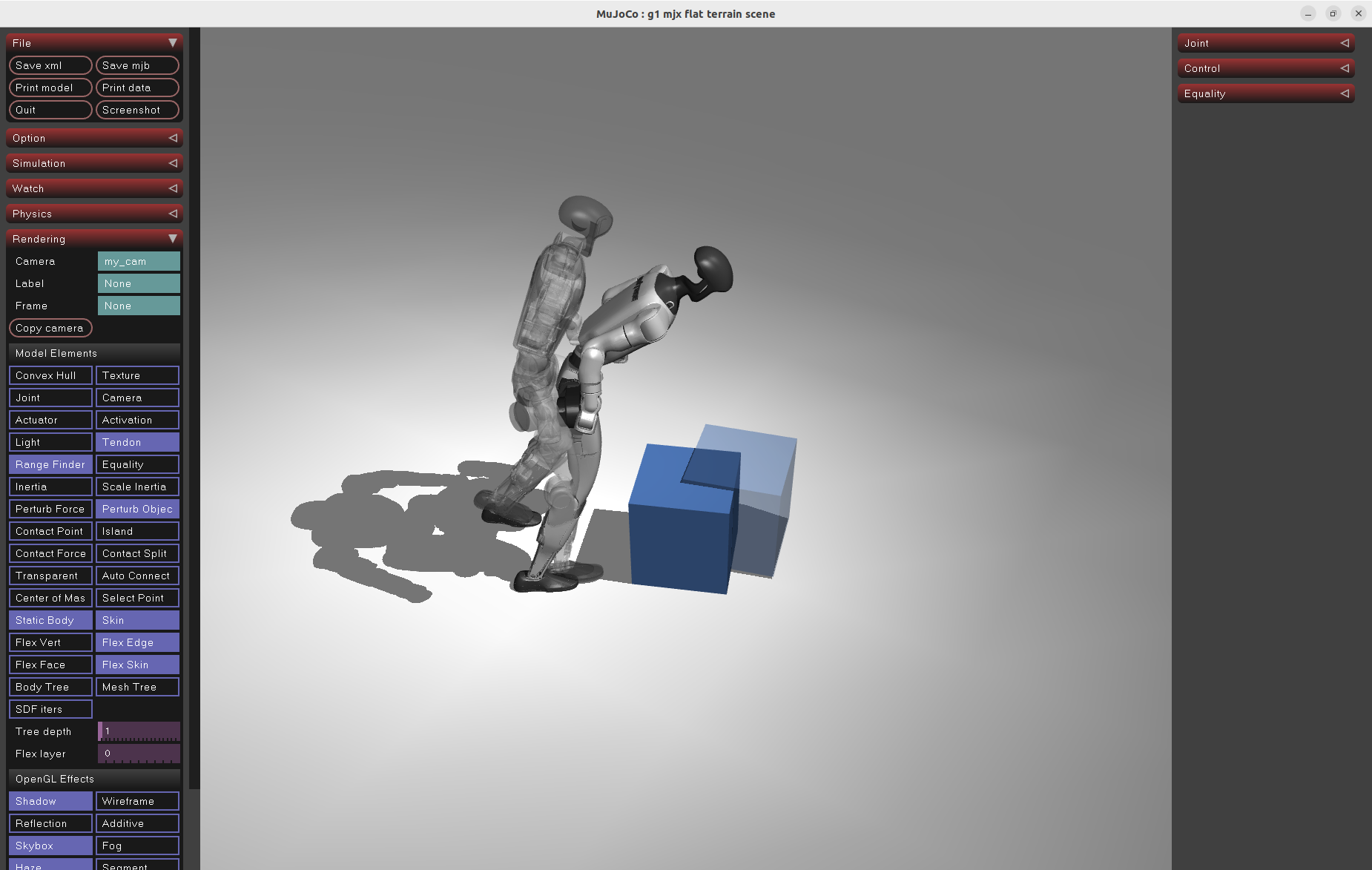}
    }
    \hspace{-3mm}
    \subfloat[SBTO, $k=4$]{
        \includegraphics[width=0.3\linewidth, trim={21cm 12cm 23cm 7cm},clip]{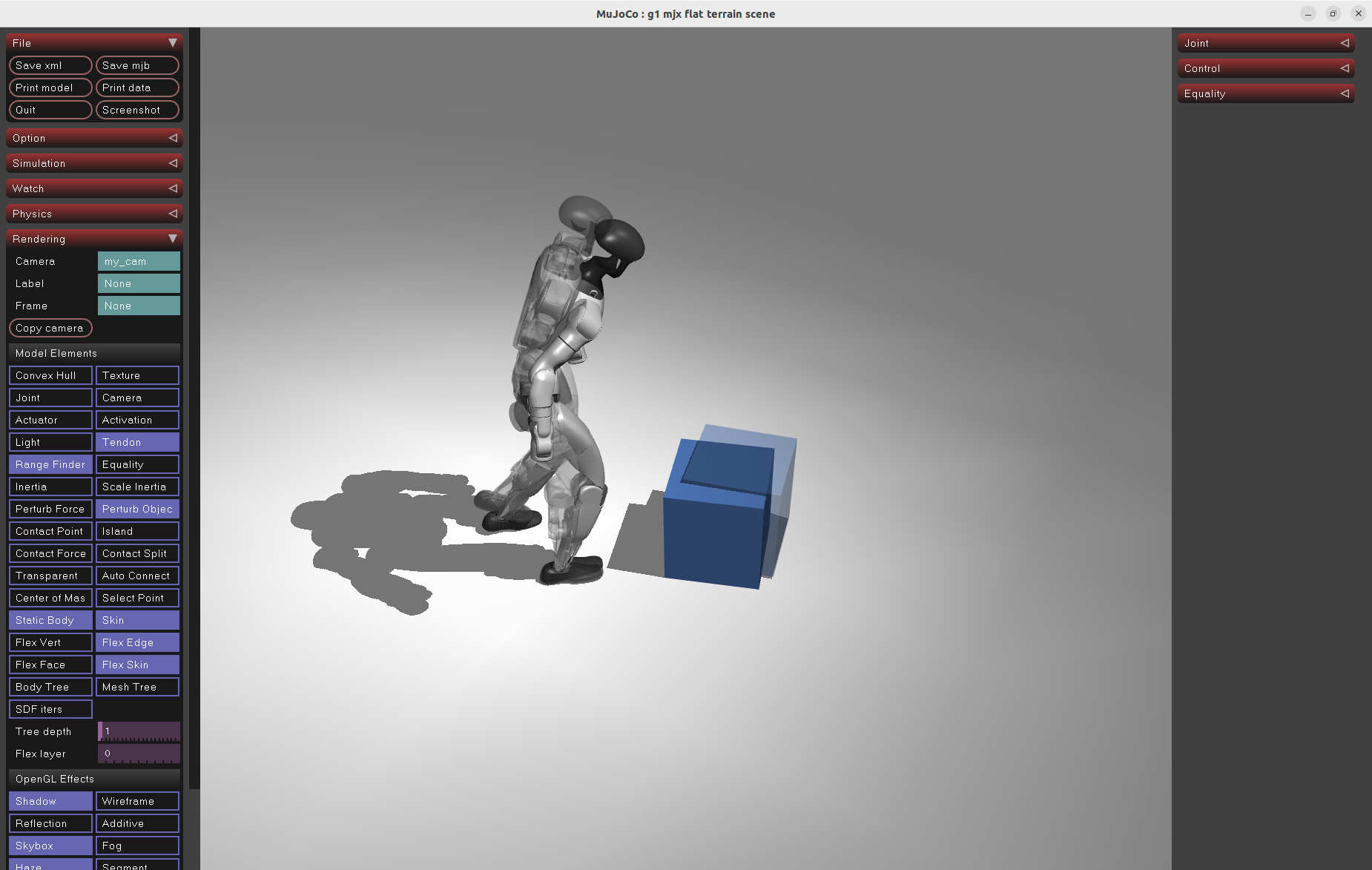}
    }
    \hspace{-3mm}
    \subfloat[SBTO, $k=7$]{
        \includegraphics[width=0.3\linewidth, trim={21cm 12cm 23cm 7cm},clip]{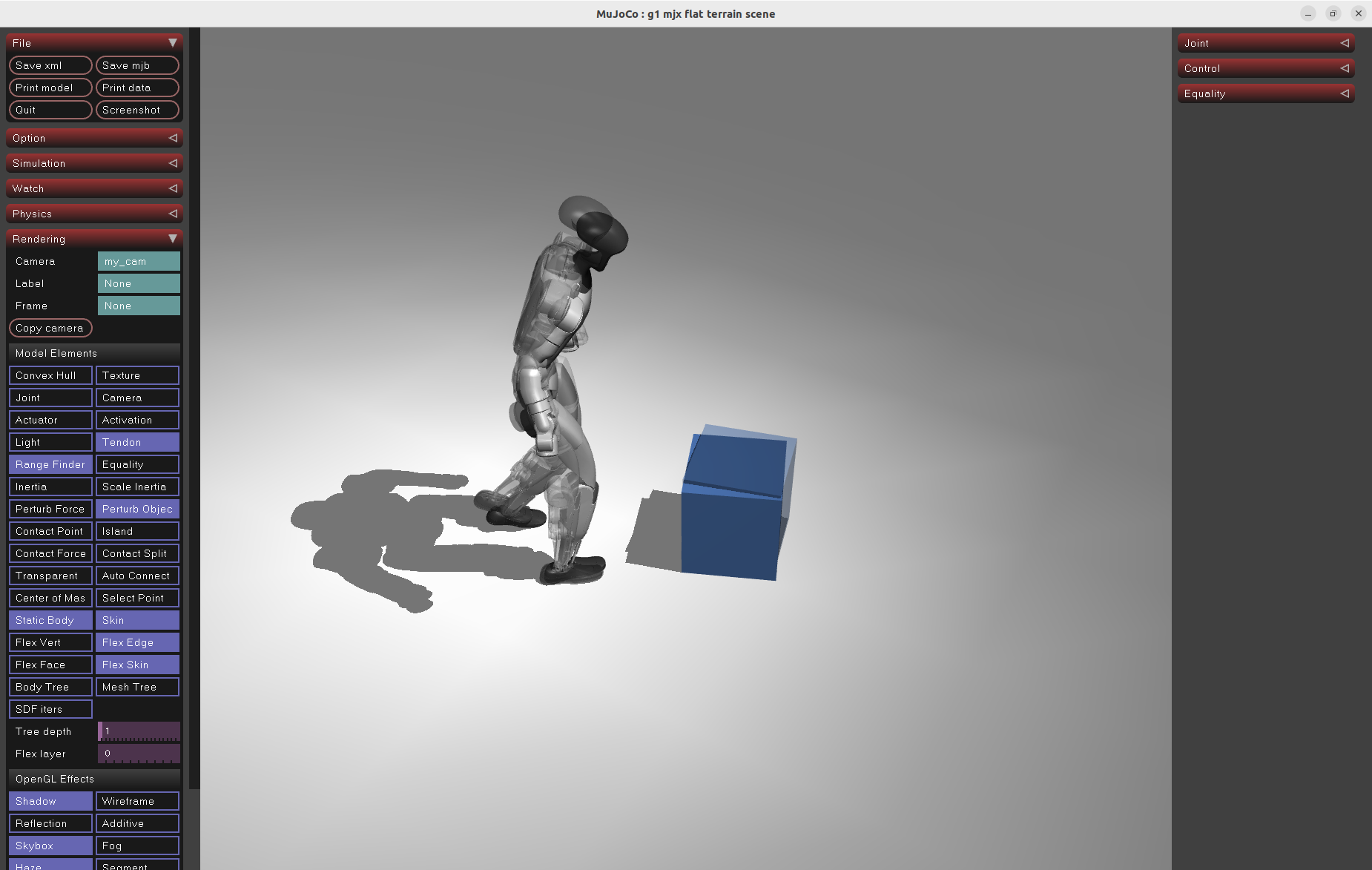}
    }
    \\
    \vspace{1mm}

    \subfloat[FHTO ($4.6$ s)]{
        \includegraphics[width=0.3\linewidth, trim={21cm 12cm 23cm 7cm},clip]{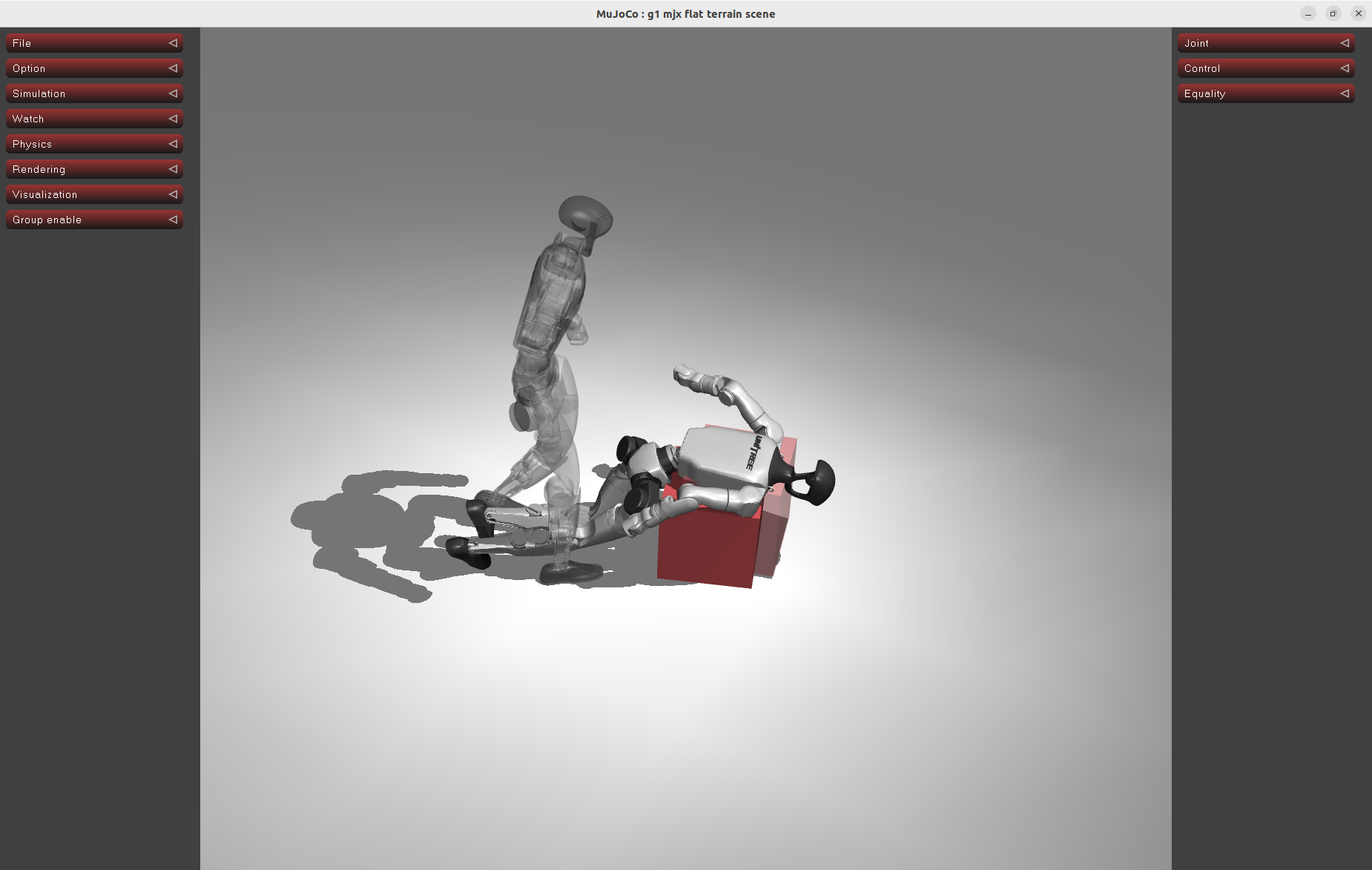}
        \label{fig:FHTO_4.6s}
    }
    \hspace{-4.2mm}
    \subfloat[FHTO ($1$ s)]{
        \includegraphics[width=0.3\linewidth, trim={21cm 12cm 23cm 7cm},clip]{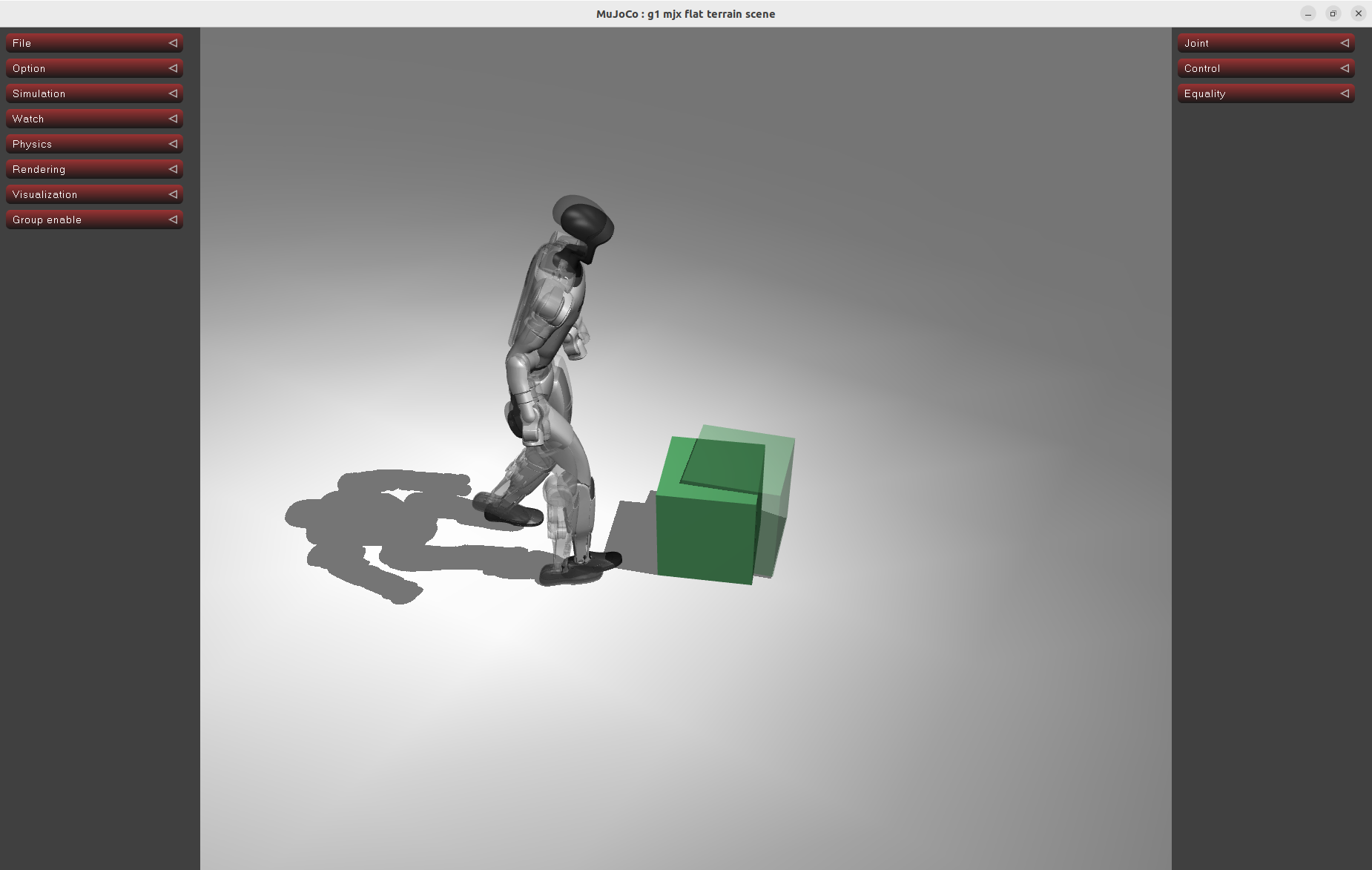}
        \label{fig:FHTO_1s}
    }
    \hspace{-4.2mm}
    \subfloat[SPIDER]{
        \includegraphics[width=0.3\linewidth, trim={21cm 12cm 23cm 7cm},clip]{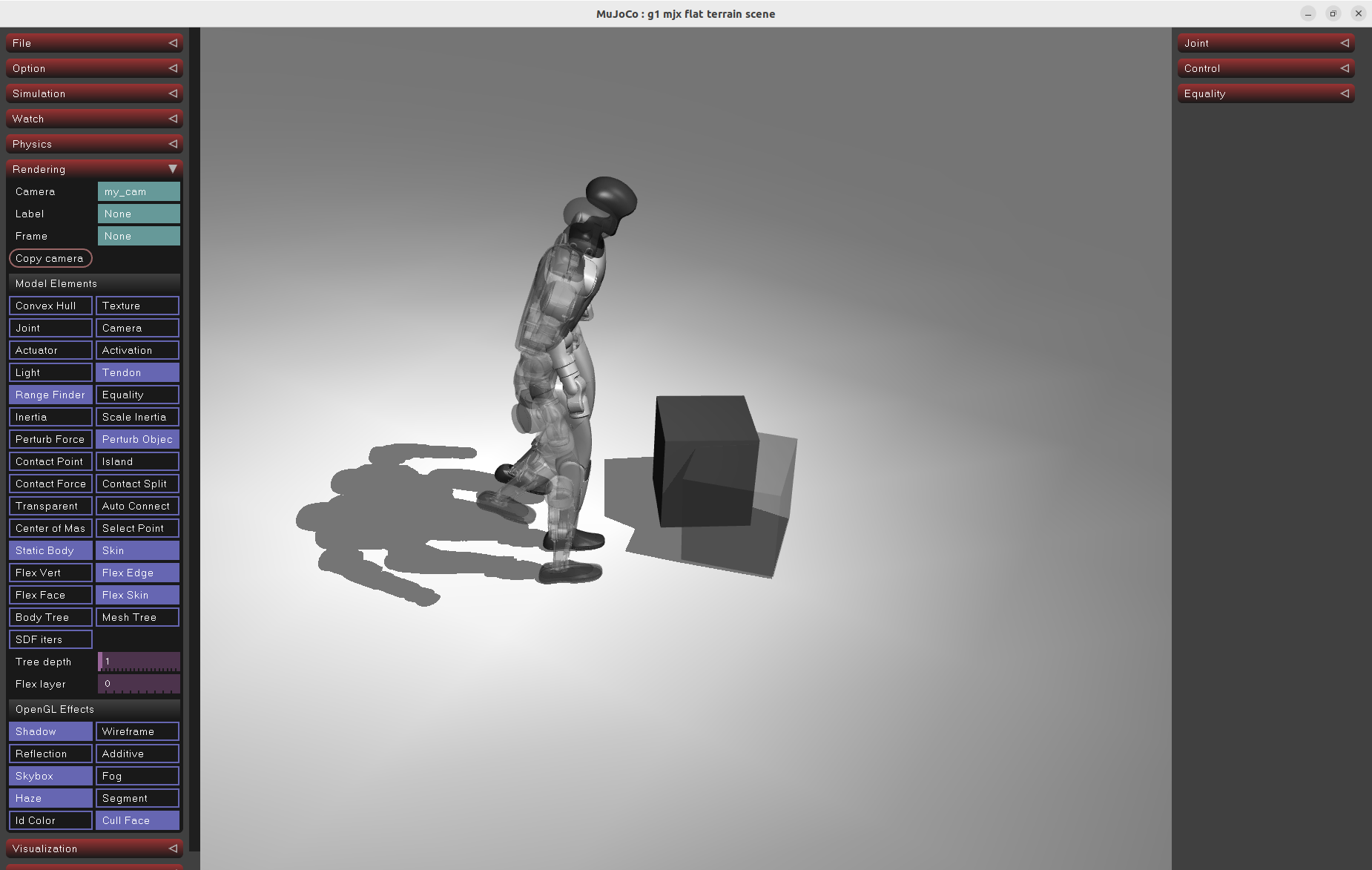}
        \label{fig:SPIDER_1s}
    }
    \vspace{-2mm}
    \caption{Trajectory snapshots at $t^0 = 1$ s for the different baselines. 
    Top row: SBTO, the box position error decreases across successive increments. 
    Bottom row: FHTO with different horizon and SPIDER baseline. 
    The reference is depicted in transparent.}
    \label{fig:snapshots_effective_horizon}
    \vspace{-3mm}
\end{figure}

\begin{figure}[t]
    \centering
    \includegraphics[width=0.95\linewidth]{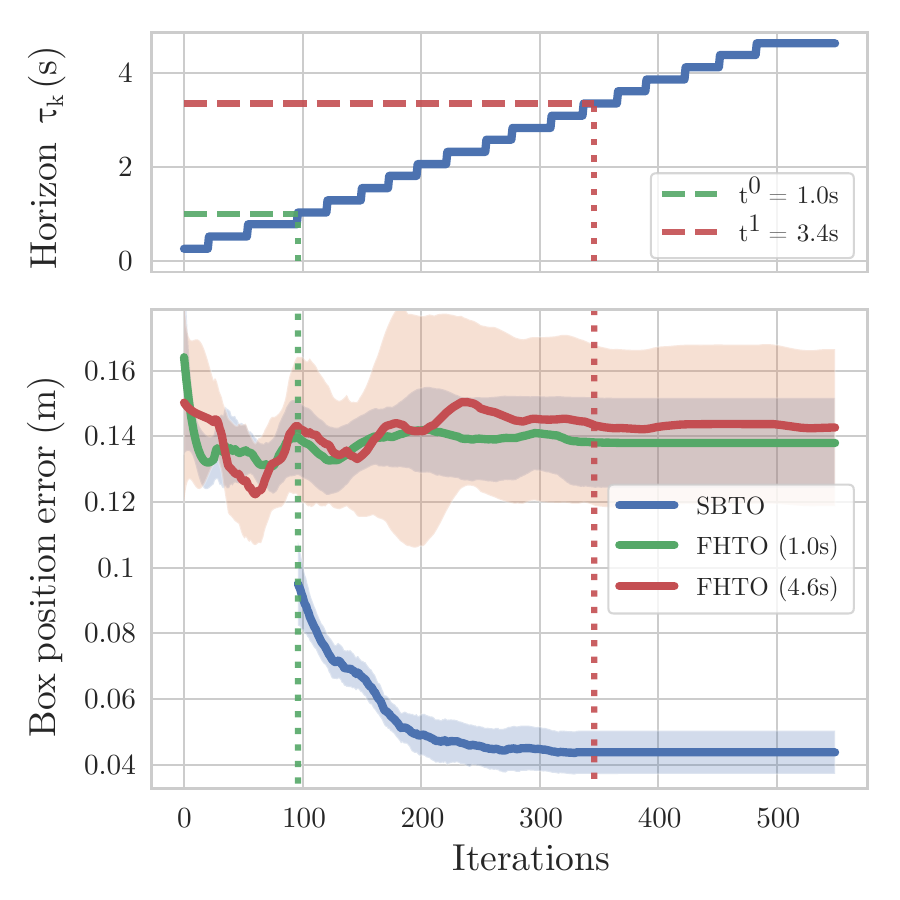}
    \vspace{-2mm}
    \caption{
    Evolution of the object position error at time $t^0$ during the optimization. The object position error steadily decreases for about $200$ iterations with SBTO. This shows that the first knots are still being optimized even after $10$ increments of the horizon, which corresponds to an effective horizon of around $3.4$ s (see vertical and horizontal red lines). Other baselines fails as the position error remains too high.
    }
    \label{fig:box_pose_error_vs_it}
    \vspace{-7mm}
\end{figure}

To qualitatively highlight these properties, we considered a specific motion, i.e., $\texttt{sub\_10\_largebox\_045}$.
In this $4.6$ s reference, a box is kicked forward at the very beginning of the trajectory. The box slides on the floor and finally stops at timestep $t^0=1$ s. Only the first two knots at $\tau_0$ and $\tau_1$ are responsible for the kicking motion.
Specifically, by tracking how the box position error \textit{at the fixed time} $t^0$ evolves across optimization iterations, we can measure how long the first decision variables remain actively optimized. Since only the first knots influence the box motion at $t^0$, improvements in the box position at $t^0$ directly indicate that the initial control variables are being refined over an extended horizon.

The results are summarized in Fig. \ref{fig:box_pose_error_vs_it} (averaged over $10$ seeds).
The top plot demonstrates how the horizon length $\tau_{k}$ grows over optimization iterations. Note that the number of iterations per increment may vary, as it depends on the rate at which the covariance $\mathbf{\Sigma}$ shrinks.

The bottom plot shows the box position error at time step $t^0$ as a function of the iterations. The position error is computed from the minimum cost state trajectory $\mathbf{x}_{0:\tau_k}^*$ at each iteration.
We compare SBTO to FHTO with different fixed horizon lengths.

For SBTO, the error can only be plotted when the growing horizon $\tau_k$ reaches $t^0=1$ s, as before that, the trajectory at $t^0$ is not even being produced by the roll-outs.
We note $i^0$ the first iteration at which $t^0=1s$ is within the optimization window (see green lines) and plot the error for SBTO starting from iteration $i^0$.
In contrast, for FHTO, we only consider horizons larger than $t^0$, therefore, the error can be computed for all optimization iterations.
Snapshot of the trajectories at $t^0=1$ s for all baselines can be seen in Fig. \ref{fig:snapshots_effective_horizon}.\newline

\subsubsection{SBTO, an incremental warm-starting process}
\label{sec:SBTO_vs_FHTO}

SBTO succeeded for all $10$ runs, whereas FHTO on the full motion length ($4.6$ s) systematically failed due to the robot falling and failing to kick the box correctly. This can be seen in Figure \ref{fig:FHTO_4.6s}.
This shows two things. First, FHTO is unlikely to converge on such a complex contact-rich task. 
Second, SBTO warm-starts the full-horizon problem efficiently, as once done incrementing, the only difference between SBTO and $4.6$ s FHTO is the state $(\boldsymbol{\mu}, \boldsymbol{\Sigma})$ from which the sampling process starts.

\subsubsection{SBTO's effective horizon}
\label{sec:effective_horizon}

The box position error of SBTO decreases steadily until iteration $i^1 \simeq 340$ (vertical red line), corresponding to a total optimization horizon of $t^1=3.4$ s (horizontal red line). This horizon is substantially longer than both the per-increment look-ahead increase ($0.25$ s) and the fixed horizon used in SPIDER ($1.2$ s). We refer to  $t^1$ as the \textit{effective horizon} of SBTO.

This behavior indicates that early control variables continue to be refined over many increments (approximately $10$), meaning that the first decision variables are optimized with a tracking objective evaluated over $3$ s of future motion.

To emphasize that optimizing over a longer horizon is beneficial, we compare SBTO with a $t^0=1$ s FHTO, representative of an SBMPC-style setup. As one can see on the bottom plot of Figure \ref{fig:box_pose_error_vs_it}, FHTO fails as the final box position error remains above $10$cm. SPIDER fails for the same reason on this task.
Interestingly, the joint tracking with $1$ s FHTO seems satisfying, as one can see in Fig. \ref{fig:FHTO_1s}. A plausible explanation for the failure could be that the object position cost is high over such a short horizon.
In contrast, over longer horizons, the cumulative box position cost can only grow (since the box is not being moved after $t^0$), which effectively increases its weight in the objective.

\begin{figure}[h!]
    \centering
    \vspace{-4mm}
    \includegraphics[width=1\linewidth, trim={0 0 0 0},clip]{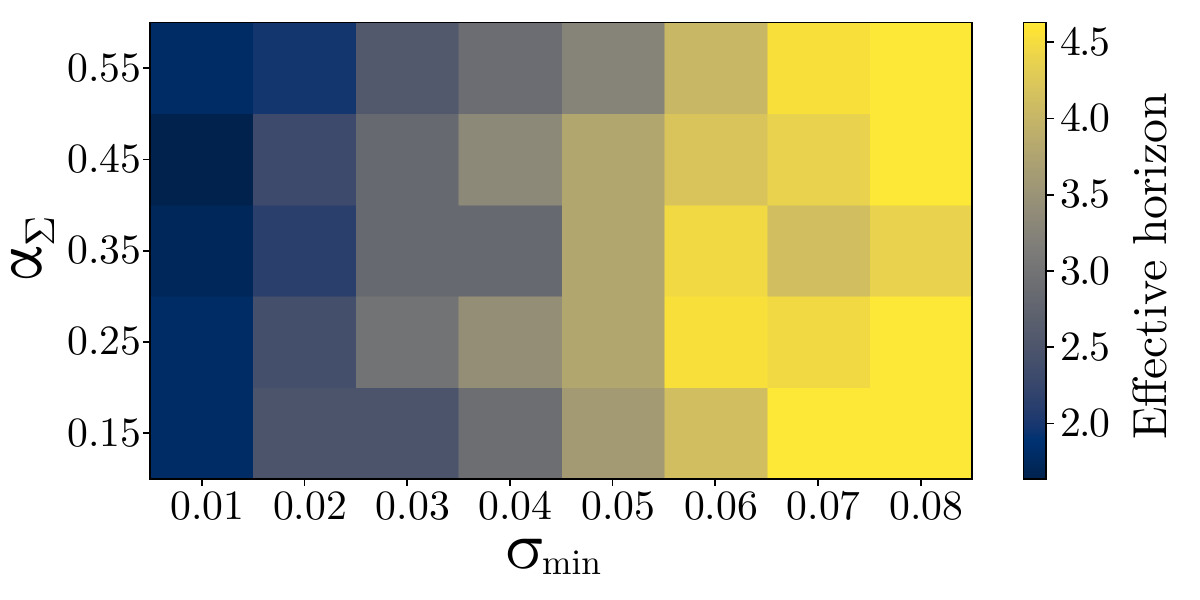}
    \vspace{-7mm}
    \caption{Effective horizon of SBTO for a parameter sweep over $\sigma_{min}$ and $\alpha_\Sigma$, averaged over $3$ runs. The effective horizon increases column by column, as $\sigma_{min}$ increases, whereas it stays almost identical for different $\alpha_\Sigma$ values.}
    \label{fig:SBTO_sweep}
    \vspace{-5mm}
\end{figure}

By performing a parameter sweep over the two main hyperparameters impacting the convergence dynamics ($\sigma_{min}$, which controls when to increment, and $\alpha_{\Sigma}$, which controls the distribution shrinking rate), we show that the effective horizon is likely not an emergent property, but is primarily governed by $\sigma_{min}$. Results are shown in Fig. \ref{fig:SBTO_sweep}.






\subsection{Demonstration Augmentation}
\label{sec:demonstration_augmentation}


SBTO produces trajectories that deviate from the kinematic reference to ensure dynamic feasibility. One way to quantify how much it could deviate is to evaluate refinement performance under changes in object properties, such as mass, size, and geometry. This evaluation is also important in practice, as collecting new demonstrations is expensive.

We evaluate the success rate of SBTO on a box with different masses and sizes, and geometries. All experiments are based on a single motion reference ($\texttt{sub10\_largebox\_084\_original}$). The original object size is a cubic box of length $0.31$cm and $0.6$kg mass. We use the same cost terms and optimization settings as in previous experiments.

\begin{figure}[h!]
    \centering

    \subfloat{
        \includegraphics[width=0.45\linewidth, trim={15cm 8cm 15cm 5cm},clip]{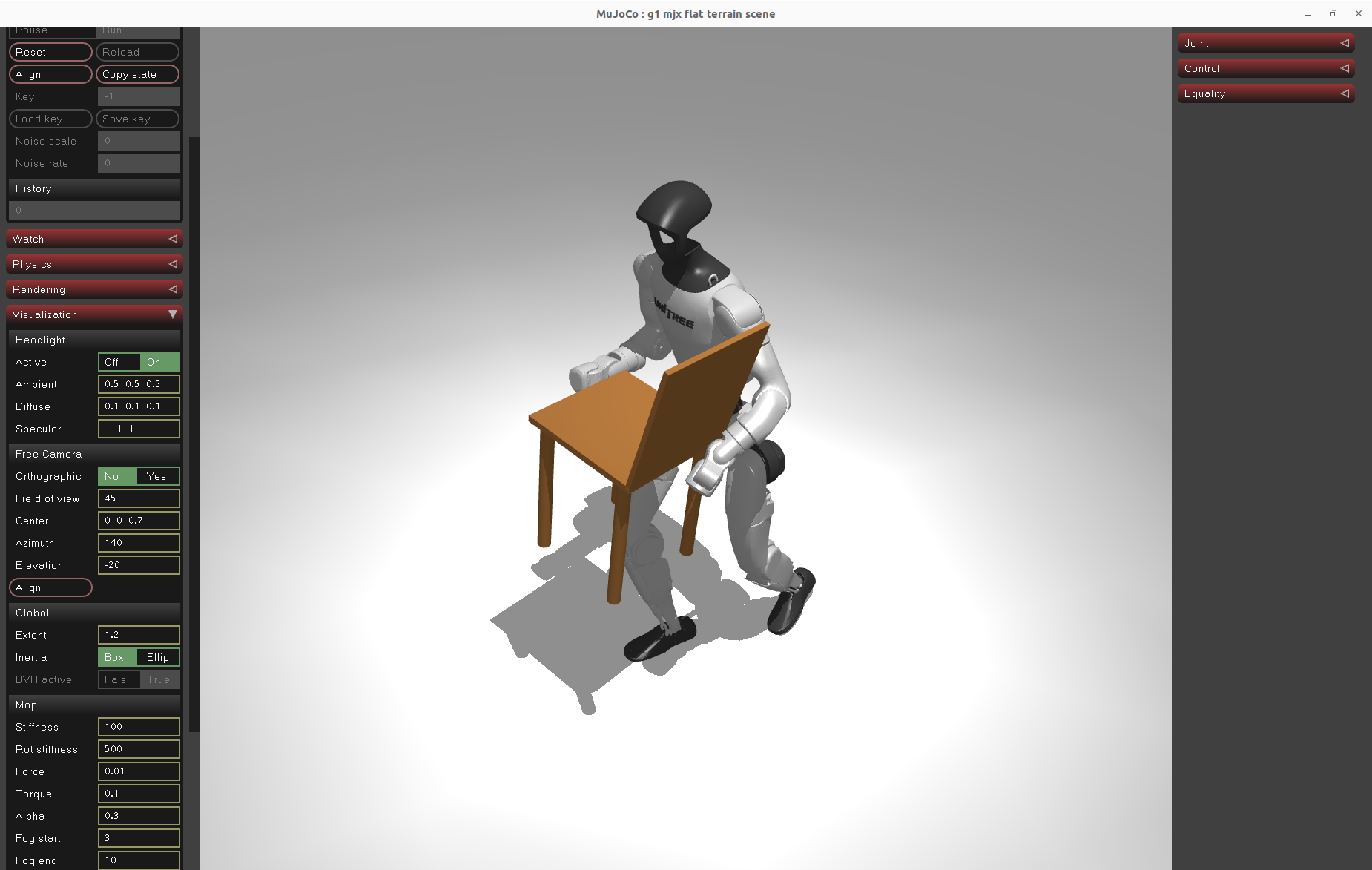}
    }
    \hspace{-1mm}
    \subfloat{
        \includegraphics[width=0.45\linewidth, trim={15cm 8cm 15cm 5cm},clip]{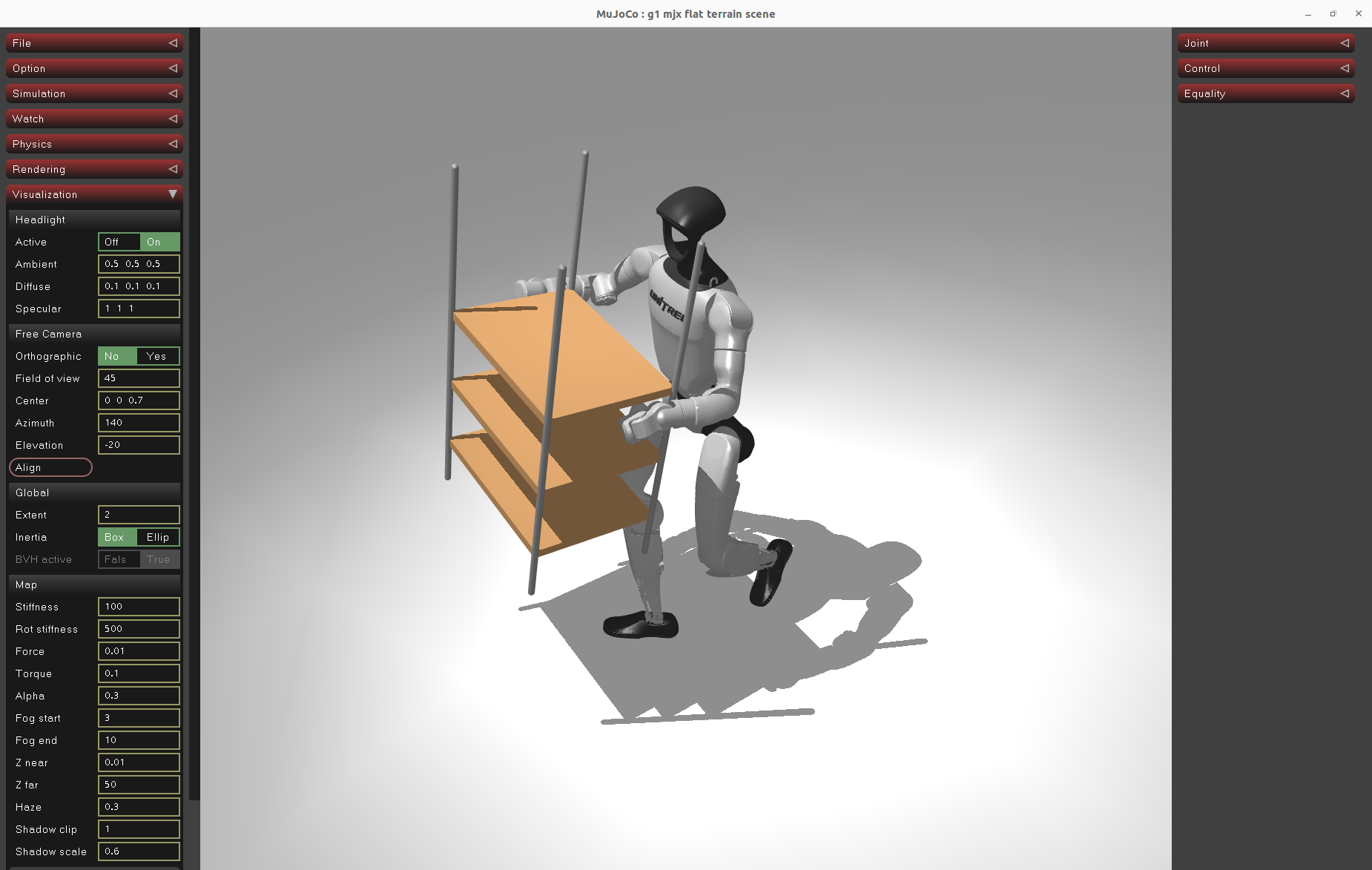}
    }

    \caption{Trajectory snapshots of \texttt{sub\_10\_largebox\_084} with the original box geometry being replaced by a chair (left) and a shelf (right).}
    \label{fig:demonstration_augmentation}
    \vspace{-5mm}
\end{figure}

SBTO successfully handled boxes with masses ranging from $0.1$ to $8$kg and sizes ranging from $0.2$m to $0.4$m.
Furthermore, the dynamic refinement process generalized beyond boxes and was also successful on a cylinder (diameter and height of $0.31$m), a chair, and a shelf (as can be seen on Figure \ref{fig:demonstration_augmentation}). This shows that a single demonstration can be refined into dynamically feasible motions across a diverse set of object geometries and physical properties.

\begin{table}[h]
\centering
\caption{Reward terms used by the RL tracking controller. }
\vspace{-2mm}
\label{tab:reward_terms}
\small
\setlength{\tabcolsep}{4pt}
\renewcommand{\arraystretch}{1.1}
\begin{tabular}{@{}lcc@{}}
\toprule
\textbf{Reward Term} & \textbf{Equation} & \textbf{Weight} \\ 
\midrule
\multicolumn{3}{l}{\textit{Motion Tracking}} \\
\midrule
Root Position &
$\exp\!\left(-5\|\mathbf{p}_t - \mathbf{p}_t^{\text{ref}}\|^2\right)$ & 0.5 \\
Root Orientation &
$\exp\!\left(-3\|\mathbf{q}_t - \mathbf{q}_t^{\text{ref}}\|^2\right)$ & 0.5 \\
Body Position &
$\exp\!\left(-5\|\mathbf{p}_{\text{body},t} - \mathbf{p}_{\text{body},t}^{\text{ref}}\|^2\right)$ & 1.0 \\
Body Orientation &
$\exp\!\left(
-3\|\boldsymbol{\theta}_{\text{body},t}
- \boldsymbol{\theta}_{\text{body},t}^{\text{ref}}\|^2
\right)$ & 1.0 \\
Body Linear Velocity &
$\exp\!\left(-0.5\|\mathbf{v}_t - \mathbf{v}_t^{\text{ref}}\|^2\right)$ & 1.0 \\
Body Angular Velocity &
$\exp\!\left(-0.05\|\boldsymbol{\omega}_t - \boldsymbol{\omega}_t^{\text{ref}}\|^2\right)$ & 1.0 \\
Joint Pos Tracking &
$\exp\!\left(-5\|\boldsymbol{u}_t - \boldsymbol{u}_t^{\text{ref}}\|^2\right)$ & 2.0 \\
\midrule
\multicolumn{3}{l}{\textit{Object Tracking}} \\
\midrule
Contact Match &
\scalebox{0.85}{$\mathbb{I}(c = c^{\text{ref}})\, \cdot
\exp\!\left(-0.1\left(\left\|\boldsymbol{F}_t\right\| - 10\right)\right)$} & 1.25 \\
Object Position &
$\exp\!\left(-8\|\mathbf{p}_{\text{obj},t} - \mathbf{p}_{\text{obj},t}^{\text{ref}}\|^2\right)$ & 1.0 \\
Object Orientation &
$\exp\!\left(-5\|\boldsymbol{\theta}_{\text{obj},t} - \boldsymbol{\theta}_{\text{obj},t}^{\text{ref}}\|^2\right)$ & 1.0 \\
Object Linear Velocity &
$\exp\!\left(-2\|\mathbf{v}_{\text{obj},t} - \mathbf{v}_{\text{obj},t}^{\text{ref}}\|^2\right)$ & 1.0 \\
Object Angular Velocity &
$\exp\!\left(-0.2\|\boldsymbol{\omega}_{\text{obj},t} - \boldsymbol{\omega}_{\text{obj},t}^{\text{ref}}\|^2\right)$ & 1.0 \\
\midrule
\multicolumn{3}{l}{\textit{Regularization}} \\
\midrule
Action Rate &
$-\|\mathbf{a}_t - \mathbf{a}_{t-1}\|^2$ & $-0.1$ \\
Joint Limit &
$-\sum_i \phi(q_i, q_{i,\min}, q_{i,\max})$ & $-10.0$ \\
Self-collisions &
$\sum_{c \in \mathcal{C}_{\text{self}}} 1$ & $-1.0$ \\
\bottomrule
\end{tabular}
\vspace{-5mm}
\end{table}

\subsection{Motion Tracking using Reinforcement Learning}
\label{sec:rl_eval}

With access to physically consistent trajectories for humanoid-object loco-manipulation, we conduct extensive experiments on training RL tracking controllers using PPO~\cite{liao2025beyondmimicmotiontrackingversatile}, while using a residual action space as in~\cite{weng2025hdmilearninginteractivehumanoid}. We use additional observations to track the object trajectory throughout the episode. In addition to the one-step desired robot trajectory, the policy observes the one-step object pose error with respect to the desired pose, as well as the object pose expressed in the robot frame. As an additional objective for our controller, we also task the policy to track the object pose. We use contact rewards to incentivize contact for the different end-effectors with the object. The contact reward also gradually penalizes forces that exceed $10$N. Since we have access to physically consistent trajectories using our method, we can naturally extract accurate contacts from the same simulator, without relying on any heuristics. For all of our experiments, we used the same set of reward weights as illustrated in Table~\ref{tab:reward_terms}.

\begin{table*}[htbp]
\centering
\caption{Downstream RL policy evaluation using different references to track}
\vspace{-2mm}
\begin{tabular}{lccccc}
\hline
\textbf{Method} &
\textbf{Success Rate} $\uparrow$ &
\textbf{MPKPE} $\downarrow$ &
\textbf{Object Pos Error} $\downarrow$ &
\textbf{Object Ori Error} $\downarrow$ \\
 & \textbf{(\%)} & \textbf{(cm)} & \textbf{(cm)} & \textbf{(rad)} \\
\hline
OmniRetarget \cite{yang2025omniretargetinteractionpreservingdatageneration} &
$79.41 \pm 32.57$ &
$3.67 \pm 0.33$ &
$12.50 \pm 6.70$ &
$0.18 \pm 0.09$ \\
\textbf{DynaRetarget} (Ours) &
$\mathbf{97.09 \pm 2.31}$ &
$\mathbf{3.57 \pm 0.46}$ &
$\mathbf{8.81 \pm 1.16}$ &
$\mathbf{0.11 \pm 0.02}$ \\
\hline
\end{tabular}
\label{tab:rl_comparison}
\vspace{-5mm}
\end{table*}

To enable robust transfer of our loco-manipulation policies to the real robot, we employ additional domain randomization techniques apart from those used in~\cite{liao2025beyondmimicmotiontrackingversatile} upon resetting the episode with adaptive sampling, which samples more difficult portions of the trajectory more frequently than others. Specifically, when the episode is reset during the initial frames, we randomize the object pose $(\mathbf{p}_{\text{obj},t}, \boldsymbol{\theta}_{\text{obj},t})$ around the desired object pose $(\mathbf{p}_{\text{obj},t}^{\text{ref}}, \boldsymbol{\theta}_{\text{obj},t}^{\text{ref}})$. When the reset occurs after the initial frames, we instead randomize the object velocities $(\mathbf{v}_{\text{obj},t}, \boldsymbol{\omega}_{\text{obj},t})$ around their desired values $(\mathbf{v}_{\text{obj},t}^{\text{ref}}, \boldsymbol{\omega}_{\text{obj},t}^{\text{ref}})$. In addition, we apply random external pushes to the object at random intervals during the episode, randomize the object's friction parameters, and vary its mass around the nominal value. Crucially, we do not introduce any additional termination conditions related to object tracking, as we found that such terminations consistently degrade performance and exacerbate the learning difficulty when combined with adaptive sampling during training from scratch.
We use \texttt{mjlab}, which uses the GPU-optimized version of MuJoCo, with IsaacLab-style~\cite{mittal2025isaaclab} API design for training our motion tracking controllers with $8192$ envs for $10000$ iterations on a single NVIDIA RTX 4090 GPU.

To highlight the importance of physical consistency of the trajectories for downstream RL policy performance, we compare success rates and tracking metrics against policies trained with trajectories from OmniRetarget \cite{yang2025omniretargetinteractionpreservingdatageneration} in Table~\ref{tab:rl_comparison}. We left out comparisons with SBMPC-based methods since they do not succeed on diverse enough motions for reasons explicated in Section~\ref{subsec:sbmpc-issues} and evaluated in Table~\ref{tab:algo_comparison}. Results are averaged over $1024$ episodes spanning $8$ distinct motions and diverse initial configurations. These motions cover a broad range of motions, such as lifting, pushing with hands, pushing with legs, etc. A policy roll-out is deemed successful if the object pose remains within a predefined threshold of the desired object pose at every timestep of the trajectory, since the policy may continue to imitate the robot motion even when object tracking fails in the case of infeasible trajectories. In contrast to prior approaches that rely on artificial curriculum to mitigate such artifacts \cite{zhao2025resmimicgeneralmotiontracking}, our method enables the RL tracking controller to reliably learn even the most challenging behaviors, such as object sliding and object manipulation using the robot's legs without additional curriculum shaping, as shown in Figure~\ref{fig:teaser}.

\begin{figure}[h]
    \centering
    \subfloat[Object position reward]{
        \includegraphics[width=0.49\linewidth, trim={0.5cm 1cm 20cm 3.5cm},clip]{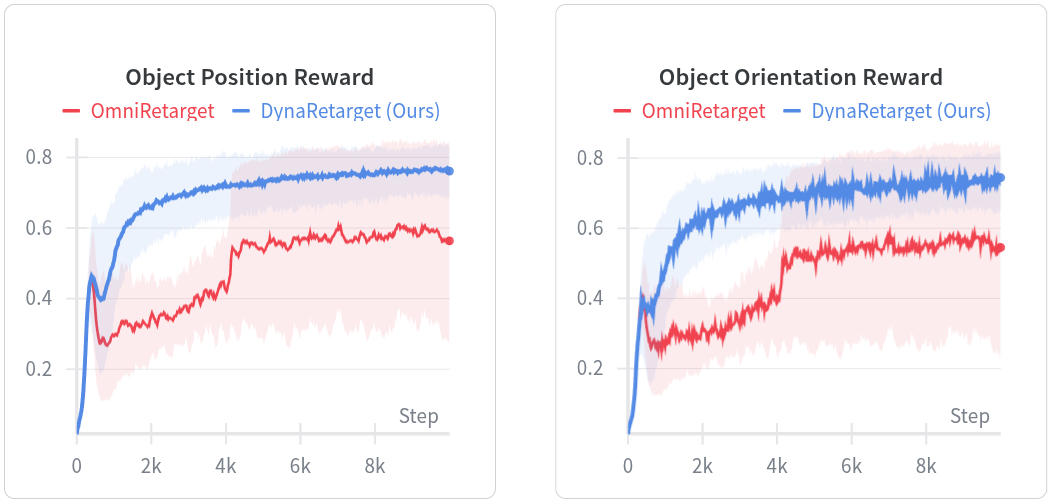}
    }
    \hspace{-5mm}
    \subfloat[Object orientation reward]{
        \includegraphics[width=0.49\linewidth, trim={20cm 1cm 1cm 3.5cm},clip]{images/rl/sample_efficiency_rl.png}
    }

    \vspace{-2mm}
     \caption{Comparison of object position and orientation tracking rewards throughout training using $3$ distinct references, from SBTO or OmniRetarget. This highlights superior performance and sample efficiency when training tracking policies on dynamically consistent trajectories.}
    \label{fig:sample_efficiency_rl}
    \vspace{-5mm}
\end{figure}

A major advantage of our method is its superior sample efficiency, as shown in Figure~\ref{fig:sample_efficiency_rl}. RL tracking policies converge significantly faster to the desired solution when trained on perfectly dynamically consistent trajectories from SBTO using the same simulator (MuJoCo), without requiring additional tuning. In contrast, policies trained on kinematically retargeted data alone either take substantially longer to converge or fail to track the object entirely, as joints have to deviate more from the reference.

\vspace{-2mm}
\section{Conclusions  and Future Work}
\label{sec:conclusion}

In this work, we presented \emph{DynaRetarget}, a complete pipeline for transferring human motion to real-world deployed control policies. The central contribution of this pipeline is SBTO, a sampling-based trajectory optimization framework that refines imperfect kinematic humanoid trajectories into dynamically feasible motions.
SBTO incrementally grows the optimization horizon, effectively warm-starting the full-horizon problem while still allowing early decision variables to be refined as the horizon grows. This strategy mitigates the convergence challenges of sampling-based methods on long-horizon problems, while simultaneously overcoming the short-sightedness of SBMPC.

We extensively evaluated SBTO on hundreds of motions and showed that it achieves a substantially higher success rate than a state-of-the-art SBMPC baseline, while producing smoother trajectories. We further demonstrated that SBTO-generated trajectories benefit downstream learning: RL tracking controllers trained on them acquire more reliable object interaction behaviors without additional curriculum shaping and transfer successfully to real hardware.

One potential way to further improve scalability is to employ a multi-modal sampling distribution instead of the current multivariate Gaussian. This would enable the simultaneous optimization of multiple candidate trajectories within a single optimization process, potentially reducing the computational cost per refined trajectory.
Another promising future direction is to use SBTO to track human keypoints directly, bypassing explicit kinematic retargeting.





\bibliographystyle{ieeetr}
\bibliography{references}

\end{document}